\documentclass[conference]{IEEEtran}
\usepackage{amsthm}
\usepackage{amssymb}
\usepackage{multirow}
\usepackage{stmaryrd}
\usepackage{booktabs}
\usepackage{tabularx}
\usepackage{caption2}

\usepackage{cite}

\ifCLASSINFOpdf
\else
  \usepackage[dvips]{graphicx}
\fi
\usepackage[cmex10]{amsmath}
\usepackage{array}

\usepackage{mdwmath}
\usepackage{mdwtab}

\usepackage{eqparbox}

\usepackage{graphicx}
\usepackage[tight,footnotesize]{subfigure}

\usepackage{stfloats}
\begin{document}
%
\title{Multi-tensor Completion for Estimating Missing Values in Video Data}



%
\author{\IEEEauthorblockN{Chao Li\IEEEauthorrefmark{1}\IEEEauthorrefmark{2},
Lili Guo\IEEEauthorrefmark{1}, and
Andrzej Cichocki\IEEEauthorrefmark{2} \emph{Fellow, IEEE}}
\IEEEauthorblockA{\IEEEauthorrefmark{1}College of Information and Communication Engineering, Harbin Engineering University, China 150001}
\IEEEauthorblockA{\IEEEauthorrefmark{2}Laboratory for Advanced Brain Signal Processing, RIKEN Brain Science Institute, Japan 351-0198\\
Email: \{chao.li, a.cichocki\}@riken.jp}
}


\maketitle

\begin{abstract}
Many tensor-based data completion methods aim to solve image and video in-painting problems. But, all methods were only developed for a single dataset. In most of real applications, we can usually obtain more than one dataset to reflect one phenomenon, and all the datasets are mutually related in some sense. Thus one question raised whether such the relationship can improve the performance of data completion or not? In the paper, we proposed a novel and efficient method by exploiting the relationship among datasets for multi-video data completion. Numerical results show that the proposed method significantly improve the performance of video in-painting, particularly in the case of very high missing percentage. 
\end{abstract}


%
\IEEEpeerreviewmaketitle

\section{Introduction}

In computer vision and graphics, many methods were developed for image and video in-painting. Since not only RGB images but also videos can be considered as multi-mode arrays, more and more attention was paid on tensor-based algorithms\cite{FaLRTC}\cite{CPWOPT}\cite{HardC}. However, it should be noted that the majority of methods cannot have a satisfactory performance when too much data was missed (e.g. the missing percentage of pixels is higher than 95\%). It is because remaining observations have been not sufficient to predict missing values well. Fortunately, in many real scenarios, more than one device is usually used for observation, and all the datasets are mutually related. Thus it is straightforward to infer that we can achieve a better performance for image and video in-painting by exploiting the relationship among multi-datasets.

In this paper, we proposed a novel method for multi-tensor completion. We individually applied a low-rank approximation on every unfolding matrix along each mode of data, and folded estimations together as the prediction of missing values. More details will be discussed in Section 2. Additionally, related work will be introduced in Section 3, and Section 4 will provide results of numerical experiments.

\section{Objective and algorithm}
Suppose that there are $K$ tensors $\underline{\mathbf{X}}_k,\,k=1,\ldots,K$ with missing values which are indexed by binary tensor $\underline{\mathbf{W}}_k,k=1,\ldots,K$ (0-unobserved, 1-observed). In order to complete datasets, for each individual tensor $\underline{\mathbf{X}}_k$, the corresponding estimation $\underline{\mathbf{Y}}_k$ can be obtained by optimizing a traditional objective as \cite{FaLRTC}\cite{HardC}
\begin{equation}
\begin{split}
&\min\hspace{3.3mm}\qquad{}f\left(\underline{\mathbf{Y}}_k\right)=\Vert\underline{\mathbf{Y}}_k\Vert_{*}\\[2mm]
&s.t.\qquad{}P_{\,\underline{\mathbf{W}}_k}\left(\underline{\mathbf{Y}}_k\right)=P_{\,\underline{\mathbf{W}}_k}\left(\underline{\mathbf{X}}_k\right)
\end{split} \label{nuclear1}
\end{equation}
where $\Vert\cdot\Vert_{*}$ denotes nuclear norm of a tensor, and operator $P_{\underline{\mathbf{W}}_k}\left(\cdot\right)$ denotes choosing observed elements of which the corresponding elements in $\underline{\mathbf{W}}_k$ are equal to 1. In multi-tensor case, and by the definition of the unclear norm of a tensor in \cite{NN}, we have
\begin{equation}
\begin{split}
&\min\qquad{}f\left(\underline{\mathbf{Y}}_1,\ldots,\underline{\mathbf{Y}}_K\right)=\sum_{k=1}^K\sum_{l=1}^{L_k}\alpha_{k,l}\Vert\mathbf{Y}_k^{\left(l\right)}\Vert_{*} \\[2mm]
&s.t.\hspace{1mm}\qquad{}P_{\,\underline{\mathbf{W}}_k}\left(\underline{\mathbf{Y}}_k\right)=P_{\,\underline{\mathbf{W}}_k}\left(\underline{\mathbf{X}}_k\right),\qquad{}k=1,\ldots,K
\end{split} \label{nuclear2}
\end{equation}
where $\mathbf{Y}_k^{\left(l\right)}$ denotes unfolding tensor $\underline{\mathbf{Y}}_k$ along the $l$th mode, $L_k,\,k=1,\ldots,K$ is equal to the number of modes of $\underline{\mathbf{X}}_k$, and $\alpha_{k,l},\,k=1,\ldots,K,\,l=1,\ldots,L_k$ are weights satisfying $\sum_l^{L_k}\alpha_{k,l}=1,\,\forall{}k$. In order to search the optimal point of Eq.\ref{nuclear2} efficiently, we use a low-rank factorization with Frobenius norm regularization to approximate each unfolding of tensors. Specifically, we have an equivalent form of Eq.\ref{nuclear2} as
\begin{equation}
\begin{split}
&\min\,f\left(\left\{\underline{\mathbf{Y}}_k,\mathbf{U}_{k,l},\mathbf{V}_{k,l}\vert\forall{}k,l\right\}\right)\\[2mm]
&\hspace{10mm}=\sum_{k=1}^K\sum_{l=1}^{L_k}\alpha_{k,l}\Vert\mathbf{Y}_k^{\left(l\right)}-\mathbf{U}_{k,l}\mathbf{V}_{k,l}^\mathrm{T}\Vert_F^2\\[2mm]
&\hspace{10mm}+\lambda\Vert\mathbf{U}_{k,l}\Vert_F^2+\lambda\Vert\mathbf{V}_{k,l}\Vert_F^2 \\[5mm]
&s.t.\hspace{2mm}\,P_{\,\underline{\mathbf{W}}_k}\left(\underline{\mathbf{Y}}_k\right)=P_{\,\underline{\mathbf{W}}_k}\left(\underline{\mathbf{X}}_k\right),\qquad{}k=1,\ldots,K
\end{split} \label{Objective}
\end{equation}
where $\lambda>0$ is tuning parameter, and $\mathbf{U}_{k,l},\mathbf{V}_{k,l},\,\forall{}k,l$ denotes factor matrices for $\mathbf{Y}_k^{\left(l\right)},\,\forall{}k,l$. It is well known that Frobenius norm regularization on each factor matrices $\mathbf{U}_{k,l},\mathbf{V}_{k,l},\,\forall{}k,l$ results in a low-rank approximation of $\mathbf{Y}_k^{\left(l\right)},\,\forall{}k,l$\cite{MMMF}. Thus Eq.\ref{Objective} can be considered as a equivalent form of Eq.\ref{nuclear2}.

For related multi-tensor, it is straightforward to infer that datasets would share information along some modes. Thus we suppose that there exists $k_1\neq{}k_2$, or $l_1\neq{}l_2$, such that $\mathbf{U}_{k_1,l_1}=\mathbf{U}_{k_2,l_2}$. It will be found that this assumption plays the key role to improve the performance for completion. To search the local optimal point of Eq.\ref{Objective}, we can alternately update $\mathbf{U}_{k_0,l_0},\mathbf{V}_{k_0,l_0},\underline{\mathbf{Y}}_{k_0},\,\forall{}k_0,l_0$ by
\newpage
\begin{align}
&\mathbf{U}_{k_0,l_0}&\gets&\left(\sum_{\begin{subarray}{c}\forall{}k,l,\,\mathrm{s.t.}\\\mathbf{U}_{k,l}=\mathbf{U}_{k_0,l_0}\end{subarray}}\alpha_{k,l}\mathbf{Y}_k^{\left(l\right)}\mathbf{V}_{k,l}\right)\cdot\nonumber\\
&&&\left(\sum_{\begin{subarray}{c}\forall{}k,l,\,\mathrm{s.t.}\\\mathbf{U}_{k,l}=\mathbf{U}_{k_0,l_0}\end{subarray}}\alpha_{k,l}\mathbf{V}_{k,l}^\mathrm{T}\mathbf{V}_{k,l}+\lambda\mathbf{I}\right)^{-1},\forall{}k_0,l_0 \label{update_U} \\
&\mathbf{V}_{k_0,l_0}&\gets&\left(\alpha_{k_0,l_0}\mathbf{Y}_{k_0}^{\left(l_0\right)\mathrm{T}}\mathbf{U}_{k_0,l_0}\right)\cdot\nonumber\\
&&&\left(\alpha_{k_0,l_0}\mathbf{U}_{k_0,l_0}^\mathrm{T}\mathbf{U}_{k_0,l_0}+\lambda\mathbf{I}\right)^{-1},\,\forall{}k_0,l_0 \\[2mm]
&\underline{\mathbf{Y}}_{k_0}&\gets&\sum_{l=1}^{L_{k_0}}\alpha_{k_0,l}\mathrm{ten}_l\left(\mathbf{U}_{k_0,l}\mathbf{V}_{k_0,l}^\mathrm{T}\right)\quad\forall{}k_0 \\[2mm]
&P_{\,\underline{\mathbf{W}}_{k_0}}\left(\underline{\mathbf{Y}}_{k_0}\right)&\gets&{}P_{\,\underline{\mathbf{W}}_{k_0}}\left(\underline{\mathbf{X}}_{k_0}\right)\quad\forall{}k_0  \label{update_Y2}
\end{align}
where operator $\mathrm{ten}_l\left(\cdot\right),\,\forall{}l$ denotes transforming a matrix into a tensor which is opposite to the operator $\left(\cdot\right)^{\left(l\right)},\,\forall{}l$, and $\mathbf{I}$ denotes a identity matrix. In the algorithm, Eq.\ref{update_U}-\ref{update_Y2} are alternately used to update $\mathbf{U}_{k,l},\mathbf{V}_{k,l},\underline{\mathbf{Y}}_k,\,\forall{}k,l$ until convergence. For initialization, choosing both $\mathbf{U}_{k,l},\mathbf{V}_{k,l},\,\forall{}k,l$ randomly and by Singular Value Decomposition (SVD) are recommended, and missing elements in $\underline{\mathbf{Y}}_k,\,\forall{}k$ can be initialized by zero. Compared to other traditional nuclear norm based methods, our method is non-convex. But it can stably provide a good performance in practice.
\section{Related work}
Many tensor based methods for completion have been developed\cite{CPWOPT}\cite{FaLRTC}\cite{HardC}\cite{TMac}\cite{BayCP}, where \cite{CPWOPT} and \cite{BayCP} are based on tensor decomposition to predict missing values while \cite{FaLRTC}\cite{HardC}\cite{TMac} minimize nuclear norm of a tensor directly. But all of them just focus on a single tensor data. It is worthwhile to notice that TMac proposed in \cite{TMac} used a similar optimization method to our method. However, we use the Frobenius norm regularization on factor matrices to control the rank of unfolding of tensors but TMac only justifies the number of column of latent matrices to control the nuclear norm.

In many different field, multi-data analysis or group analysis was also deeply discussed\cite{CIFA}\cite{GCMF}\cite{cCMF}. But the majority of methods focus on common and distinctive component extraction. In this paper, we do not care latent features in each dataset. Thus we can reduce the complexity of algorithms compared to tensor decomposition based method like \cite{CPWOPT}\cite{BayCP}. A multi-matrix completion method Convex Collective Matrix Factorization (cCMF) was proposed in \cite{cCMF}.  Compare to cCMF, The proposed method in this paper can be considered as an extension of cCMF into tensor problem.
\section{Experiments}
In this section, we use two experiments to evaluate the performance of the proposed method. In the first experiment, videos from IXMAS\footnote{http://4drepository.inrialpes.fr/public/viewgroup/6} were used to jointly predict missing values. IXMAS consists of videos in which 13 daily-live motions (e.g. sit down, check watch, kick, etc.) performed by 11 actors, actors can choose freely position and orientation, and each action was simultaneously recorded by 5 cameras. In the experiment, we choose 4 videos from two actors, they did the similar action (sit down), and there are two views for each actor. For each video, it can be modelled as a mode-3 tensor (Pixel$\times$Channel$\times$Time) $\underline{\mathbf{X}}_k,\,k=1,2,3,4$. Since two actors did similar action and each action was recorded by two cameras, videos recorded from the same camera can be supposed to have common pixel and channel information, and videos from the same actors share time-mode information. Under such an assumption, we implement the proposed method for video in-painting. 

\begin{table*}[htbp]
\centering
\caption{\label{T_1}RSE of prediction of multi-video in-painting in different missing percentage with 10 runs}
\begin{tabularx}{0.5\textwidth}{lcccccc}
\toprule\\
\multirow{3}*{Methods}&\multicolumn{2}{c}{95\%}&\multicolumn{2}{c}{50\%}&\multicolumn{2}{c}{10\%} \\\\\cline{2-3}\cline{4-5}\cline{6-7}\\
&Data 1&Data 2&Data 1&Data 2&Data 1&Data 2 \\
\midrule
Proposed method&\textbf{0.085}&\textbf{0.270}&\textbf{0.018}&\textbf{0.115}&\textbf{0.006}&\textbf{0.082} \\
FaLRTC&$0.116$&$0.403$&\textbf{0.016}&0.129&\textbf{0.006}&\textbf{0.078} \\
HardC&$0.118$&$0.559$&$0.023$&$0.133$&$0.020$&\textbf{0.079} \\
TMac&$0.145$&$0.331$&$0.033$&$0.280$&$0.020$&$0.271$ \\
CPWOPT&$0.145$&-&$0.054$&-&-&- \\
\bottomrule
\end{tabularx}
\end{table*}

\begin{figure}[!t]
\centering
\subfigure[IXMAS]{\includegraphics[width=2.5in]{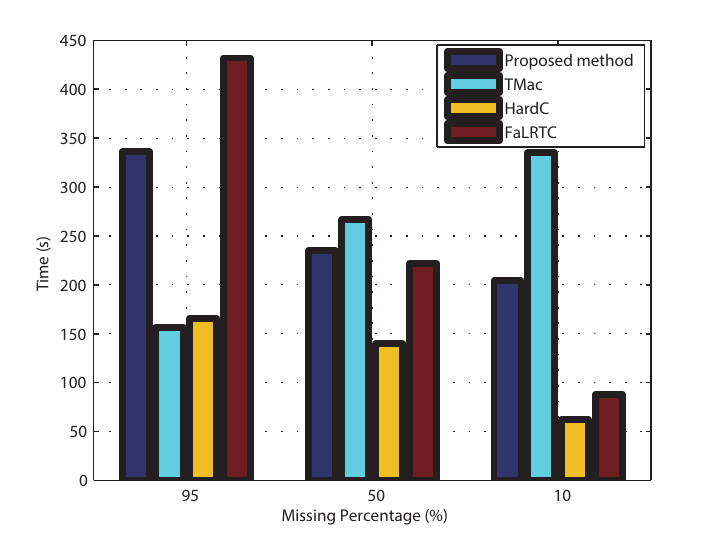}}
\hfil
\subfigure[RedSkirt]{\includegraphics[width=2.5in]{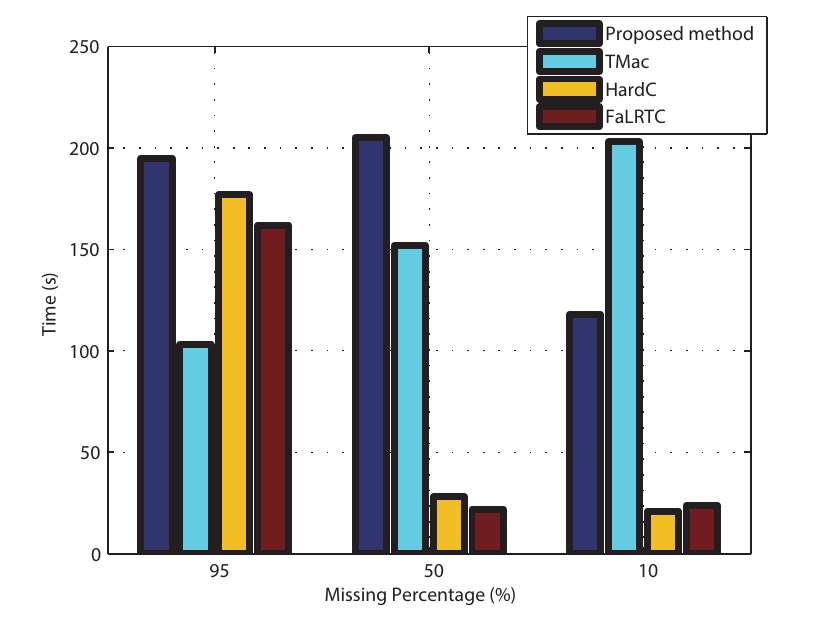}}
\caption{Average running time of algorithms in two experiments}
\label{Timeplot}
\end{figure} 

\begin{figure*}[!t]
\centering
\includegraphics[width=6in]{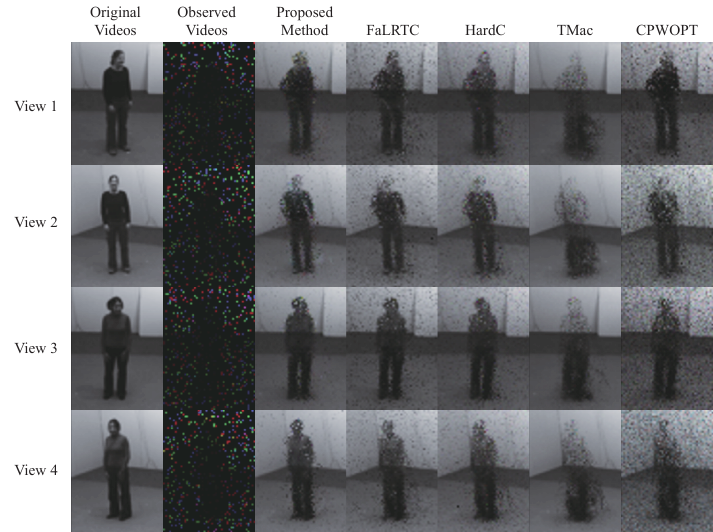}
\caption{Comparison of algorithms in the first frame of IXMAS}
\label{show}
\end{figure*}

In the second experiment, we use multi-view videos (Height$\times$Width$\times$Channel$\times$Time) of human action ( body move of a Asian lady ) which were simultaneously recorded by 3 cameras\footnote{http://media.au.tsinghua.edu.cn/index.jsp}. Since the three videos were recorded at the same time with different position, it is easy to infer that they share the information along the time mode. Further, we resize each video so that they lost common information from pixels, therefore tensor concatenation cannot be used for in-painting under this data. Such the case can be usually found when different type of cameras were used in real application. Tab.\ref{T_1} shows the Relative Square Error (RSE) for missing value prediction with different missing percentage, and Fig.\ref{Timeplot} shows comparison of the average running time from two experiments.

For comparison, FaLRTC\cite{FaLRTC}, HardC\cite{HardC}, TMac\cite{TMac}, and CPWOPT\cite{CPWOPT} was individually implemented on every video. It is shown from Tab.\ref{T_1} that the proposed method outperform other methods, particularly in the case of high missing percentage. It is because, in the case of high missing percentage, remaining information from observations for each individual video are too little for completion, but the proposed method can exploit relationship from all the datasets, and such the relationship provides more information to each dataset for missing value prediction. It is seen from Fig.\ref{Timeplot} that the running time of the proposed time is comparable to TMac, HardC, and FaLRTC, while tensor-decomposition based algorithm CPWOPT which is not shown in Fig.\ref{Timeplot} are much slower than the other four methods. Fig.\ref{show} shows completion results from all the methods to estimate the first frame of IXMAS. Fig.\ref{show} also reflects that the proposed method can achieve a better performance for video in-painting.

\section{Conclusion}
In this paper, we developed a novel completion method for multi-tensor. Compared to traditional approaches, the proposed method can exploit common information shared among datasets. By numerical results, it is demonstrated that the shared information can improve the performance in multi-videos in-painting particularly in the case of high missing percentage.


\section*{Acknowledgment}

This paper was supported by the China Scholarship Council.



%



\bibliographystyle{IEEEtran} 
\bibliography{Reference} 

\end{document}